\documentclass{IEEEtran}
\usepackage[latin9]{inputenc}
\usepackage{float}
\usepackage{amsthm}
\usepackage{amsmath}
\usepackage{amssymb}
\usepackage{graphicx}

\makeatletter

\floatstyle{ruled}
\newfloat{algorithm}{tbp}{loa}
\providecommand{\algorithmname}{Algorithm}
\floatname{algorithm}{\protect\algorithmname}

\theoremstyle{plain}
\newtheorem{thm}{\protect\theoremname}
\theoremstyle{plain}
\newtheorem{prop}[thm]{\protect\propositionname}

\usepackage{cite}\interdisplaylinepenalty=2500
\usepackage{array}

\usepackage{acronym}

\usepackage{algorithm}\usepackage{algpseudocode}

\usepackage{comment}\usepackage{bbm}\usepackage{dsfont}

\acrodef{vad}[VAD]{Voice Activity Detector}
\acrodef{GMM}[GMM]{Gaussian Mixture Model}
\acrodef{MFCC}[MFCC]{Mel-Frequency Cepstral Coefficients}
\acrodef{STFT}[STFT]{Short-Time Fourier Transform}
\acrodef{PDF}[PDF]{Probability Density Functions}
\acrodef{PSD}[PSD]{Power Spectral Density}
\acrodef{SNR}[SNR]{Signal to Noise Ratio}
\acrodef{SVM}[SVM]{Support Vector Machines}
\acrodef{omlsa}[OM-LSA]{Optimally Modified Log Spectral Amplitude}
\acrodef{MCRA}[MCRA]{Minima Controlled Recursive Averaging}
\acrodef{AUC}[AUC]{Area Under the Curve}
\acrodef{voip}[VoIP]{Voice over Internet Protocol}
\acrodef{LRT}[LRT]{Likelyhood Ratio Test}
\acrodef{DCT}[DCT]{Discrete Cosine Transform}

\@ifundefined{showcaptionsetup}{}{%
 \PassOptionsToPackage{caption=false}{subfig}}
\usepackage{subfig}
\makeatother

\providecommand{\propositionname}{Proposition}
\providecommand{\theoremname}{Theorem}

\begin{document}

\title{Kernel-based Sensor Fusion with Application to Audio-Visual Voice
Activity Detection}

\author{David~Dov{*},~\IEEEmembership{Student Member,~IEEE},~Ronen~Talmon,~\IEEEmembership{Member,~IEEE}
and ~Israel~Cohen, ~\IEEEmembership{Fellow,~IEEE\thanks{The authors are with the Department of Electrical Engineering, The Technion-Israel Institute of Technology, Haifa 32000, Israel (e-mail: davidd@tx.technion.ac.il; ronen@ee.technion.ac.il; \protect\\ icohen@ee.technion.ac.il).}}}
\maketitle
\begin{abstract}
In this paper, we address the problem of multiple view data fusion
in the presence of noise and interferences. Recent studies have approached
this problem using kernel methods, by relying particularly on a product
of kernels constructed separately for each view. From a graph theory
point of view, we analyze this fusion approach in a discrete setting.
More specifically, based on a  statistical model for the connectivity
between data points, we propose an algorithm for the selection of
the kernel bandwidth, a parameter, which, as we show, has important
implications on the robustness of this fusion approach to interferences.
Then, we consider the fusion of audio-visual speech signals measured
by a single microphone and by a video camera pointed to the face of
the speaker. Specifically, we address the task of voice activity detection,
i.e., the detection of speech and non-speech segments, in the presence
of structured interferences such as keyboard taps and office noise.
We propose an algorithm for voice activity detection based on the
audio-visual signal. Simulation results show that the proposed algorithm
outperforms competing fusion and voice activity detection approaches.
In addition, we demonstrate that a proper selection of the kernel
bandwidth indeed leads to improved performance. 
\end{abstract}

\section{Introduction}

Multiple view data fusion is the process of obtaining a unified representation
of data captured in multiple measurement systems of different types.
Data fusion has recently attracted a growing interest in the signal
processing and data analysis communities due to an extensive use of
multiple sensors in everyday devices such as computers and smartphones.
Often, data measured in multiple views is contaminated with noises
and interferences which are view specific, and fusing the views may
allow for obtaining representations of the data, which are robust
to the interferences. A challenging example which we consider in the
current work is the fusion of audio and visual recordings of a speaker.
While each view (audio or video) possibly consists of view-specific
interferences (e.g., acoustic noises or face movements), their fusion
may give rise to a robust representation of the speech.

In this paper, we use a kernel based geometric approach to address
the problem of multiple view data fusion. Classical methods, e.g.,
those presented in \cite{roweis2000nonlinear,balasubramanian2002isomap,belkin2003laplacian,donoho2003hessian,coifman2006diffusion},
represent a class of non-linear dimensionality reduction methods designed
for data measured in a single view. By learning geometric structures
of high dimensional data, these methods provide low dimensional representations
of the data via eigenvalue decomposition of an affinity kernel. The
low dimensional representations preserve the geometry of the data,
i.e., local affinities between data points, and they are successfully
used in a wide range of applications such as anomaly and target detection
and speech enhancement \cite{mishne2013multiscale,mishne2015graph,talmon2012supervised,talmon2013single}.
However, when the data is corrupted by structured interferences, the
kernel methods learn the structure of the interferences along with
the structure of the data. Therefore, the obtained low dimensional
representation retains the relations between the data and the interferences,
and, as a result, the kernel methods have a limited robustness to
the interferences.

The potential of improving the robustness of the obtained representations
to interferences by fusing data captured in multiple views, has recently
motivated researchers extending kernel-based geometric methods to
the multiple views case \cite{zhou2007spectral,blaschko2008correlational,de2010multi,kumar2011co,kumar2011co2,lin2011multiple,wang2012unsupervised,huang2012affinity,boots2012two,bronstein2013making,lindenbaum2015multiview,lederman2015learning}.
Among these studies, we mention the studies presented in \cite{de2010multi,wang2012unsupervised,lindenbaum2015multiview,lederman2015learning}
sharing similar ideas of constructing separate affinity kernels for
each view, and fusing the data by the product between the affinity
kernels. A method of particular interest in this work was presented
in \cite{lederman2015learning}, where special emphasis is given to
the robustness of the fusion process to interferences. The authors
presented a data fusion method termed alternating diffusion maps,
which is based on fusing the views by multiplying between affinity
kernels interpreted as employing separate diffusion processes on each
view in an alternating manner. By analyzing the method in the continuous
setting, it is shown that the interruptions of a certain view are
attenuated by the diffusion steps of the other views. 

The ability of kernel methods to properly learn the geometric structure
of the data is highly dependent on the selection of the kernel bandwidth,
which also has important implications on the robustness of the methods
to interferences. The kernel bandwidth, also called the scale parameter,
defines a local neighborhood such that all data points within the
neighborhood are considered similar, i.e., close to each other. It
has an intuitive interpretation by viewing the kernel methods from
the graph theory point of view, which we adopt throughout this paper.
The affinity kernel defines a graph whose nodes are the data points
and the edges are given by the affinities between the data points.
Accordingly, all data points located within a local neighborhood defined
by the kernel bandwidth are considered connected on the graph. In
the single view case, the kernel bandwidth is chosen according to
a trade-off. On the one hand, it has to be large enough keeping the
graph connected, which is a necessary condition for learning the geometry
of the data \cite{zelnik2004self,coifman2006diffusion,lafon2006data,von2007tutorial}.
On the other hand, the kernel bandwidth has to be as small as possible
so that data and interferences will not share the same local neighborhoods
\cite{keller2010audio}. In the multiple view case, the selection
of the kernel bandwidth is not addressed in the literature, and the
kernel bandwidths are naively chosen in previous studies as if the
data is measured in a single view. 

In this study, we address the fusion problem of data obtained in multiple
views. We revisit the alternating diffusion maps method and analyze
it from a different point of view than in \cite{lederman2015learning}
using a discrete setting. By adopting ideas from \cite{de2010multi,steinerberger2014filtering},
we study how connected data points on graphs of each view affect the
connectivity of the graph obtained by fusing the views via the product
of the affinity kernels. By assuming a statistical model on the connectivity
between data points in each view, we use a simple argument to show
that the kernel bandwidth of each view may be chosen such that the
graph defined on each single view is not connected. This allows us
to use significantly smaller kernel bandwidths improving the robustness
of the fusion process to interferences. Based on the introduced statistical
model, we propose an algorithm for the selection of the kernel bandwidth.
We note that throughout this paper we consider the selection of the
kernel bandwidth for the alternating diffusion maps method presented
in \cite{lederman2015learning}. However, the provided analysis and
the proposed algorithm may be extended with mild modifications to
the methods presented in \cite{de2010multi,wang2012unsupervised,lindenbaum2015multiview},
which are also based on the product between the affinity kernels of
the views.

Using the alternating diffusion mas with the new algorithm for determining
the kernel bandwidth, we address the problem of audio-visual voice
activity detection, where the goal is to detect segments of the measured
signal containing active speech. We consider a challenging setup in
which a speech signal is measured by a single microphone and a video
camera in the presence of high levels of acoustic noises and transients,
which are short term interruptions, e.g., keyboard taps and office
noise \cite{hirszhorn2012transient,dov2014voice}. In the video signal,
there exist natural mouth movements during non-speech periods which
 wrongly appear similar to speech. The alternating diffusion maps
method is particularly suitable for the fusion of the audio-visual
data in this setup since it integrates out the interferences, which
are view specific, i.e., transients measured in the microphone and
non-speech mouth movements measured by the camera. Based on the alternating
diffusion maps method, we propose a data-driven algorithm for voice
activity detection. The algorithm comprises a simple preprocessing
stage of feature extraction and does not require post-processing,
and, in contrast to the method we presented in \cite{dov2015audio},
it requires no training data. Our simulation results demonstrate improved
performance of the proposed algorithm compared both to a similar algorithm
based on a traditional selection of the kernel bandwidth and compared
to competing fusion schemes. 

The remainder of the paper is organized as follows. In Section \ref{sec:problem formulation},
we briefly review the alternating diffusion maps method. In Section
\ref{sec:Kernel-Bandwidth-Selection}, we analyze the method in a
discrete setting using tools from graph theory, and propose an algorithm
for kernel bandwidth selection. In section \ref{sec:Audio-Visual-Fusion}
we address the problem of audio-visual voice activity detection and
propose an algorithm based on the alternating diffusion maps method.
The improved performance of the proposed algorithm is demonstrated
in Section \ref{sec:Simulation-Results}.

\section{Review of The Alternating Diffusion Maps Method\label{sec:problem formulation}}

Consider a dataset of $N$ samples captured in two different views
given by:
\begin{equation}
\left(\mathbf{v}_{1},\mathbf{w}_{1}\right),\left(\mathbf{v}_{2},\mathbf{w}_{2}\right),...,\left(\mathbf{v}_{N},\mathbf{w}_{N}\right),\label{eq:dataset}
\end{equation}
where $\mathbf{v}_{n}\in\mathbb{R}^{L_{v}}$ and $\mathbf{w}_{n}\in\mathbb{R}^{L_{w}}$
are the $n$th data points of the first and the second views, respectively.
An example we will address under this setup is an audio-visual recording
of a speaker, where $\mathbf{v}_{n}$ is the $n$th time frame of
the signal captured in a microphone and $\mathbf{w}_{n}$ is the corresponding
video frame of the mouth region of the speaker. The alternating diffusion
maps method presented in \cite{lederman2015learning} is a kernel
based geometric method for data fusion. It is designed to reveal the
geometric structure of the data, which is mutual to the two views
ignoring the interferences, which are captured only in one of the
views. In the following, we shortly describe the construction of alternating
diffusion maps. Let $\mathbf{K}_{v}\in\mathbb{R}^{N\times N}$ be
an affinity kernel representing affinities between data points in
the first view, such that the $\left(n,m\right)$th entry of the matrix,
denoted by $K_{v}(n,m)$ is given by:
\begin{equation}
K_{v}(n,m)=\exp\left(-\frac{||\mathbf{v}_{n}-\mathbf{v}_{m}||^{2}}{\epsilon_{v}}\right),\label{eq:K_v}
\end{equation}
where $\epsilon_{v}$ is the kernel bandwidth whose selection is discussed
in details in Section \ref{sec:Kernel-Bandwidth-Selection}. The affinity
kernel $\mathbf{K}_{v}$ in \eqref{eq:K_v} defines a graph on the
dataset in the first view such that each data point is a vertex and
$K_{v}(n,m)$ is the weight of the edge between vertex $n$ and vertex
$m$. Let $\mathbf{M}_{v}\in\mathbb{R}^{N\times N}$ be a row stochastic
Markov matrix given by normalizing the rows of $\mathbf{K}_{v}$:
\begin{equation}
\mathbf{M}_{v}=\mathbf{D}_{v}^{-1}\mathbf{K}_{v},\label{eq:M_v}
\end{equation}
where $\mathbf{D}_{v}\in\mathbb{R}^{N\times N}$ is a diagonal matrix,
whose $m$th element on the diagonal is denoted by $D_{v}(m,m)$ and
is given by $D_{v}(m,m)={\displaystyle \sum_{n=1}^{N}K_{v}(n,m)}$.
In this study, we use a row normalization rather than a column normalization
used in \cite{lederman2015learning} allowing us to facilitate the
discussion and results in Section \ref{sec:Kernel-Bandwidth-Selection},
and our experimental results showed a negligible effect on the type
of the normalization. The matrix $\mathbf{M}_{v}$ defines a Markov
chain on the graph such that $\mathbf{M}_{v}(n,m)$ is the probability
of moving from data point $n$ to data point $m$ in a single step.
Similarly to $\mathbf{K}_{v}$, let $\mathbf{K}_{w}\in\mathbb{R}^{N\times N}$
be a matrix representing affinities between data points in the second
view, and let $\mathbf{M}_{w}\in\mathbb{R}^{N\times N}$ be the corresponding
row stochastic matrix. The views are fused by constructing a unified
matrix, which is denoted by $\mathbf{M}$ and is given by the product
of the row stochastic matrices \cite{lederman2015learning}: 
\begin{equation}
\mathbf{M}=\mathbf{M}_{v}\cdot\mathbf{M}_{w}.\label{eq:M}
\end{equation}
The matrix $\mathbf{M}$ is also row stochastic and it integrates
the relations between the data points over the two views; therefore,
we term it the multiple view Markov matrix. The continuous counterparts
of the matrices $\mathbf{M}_{v}$ and $\mathbf{M}_{w}$ in \eqref{eq:M}
are typically considered in the literature as diffusion operators
\cite{coifman2006diffusion}. Likewise, the authors in \cite{lederman2015learning}
considered $\mathbf{M}$ as an alternating diffusion operator consisting
of two diffusion steps on the two views, and showed that this alternating
diffusion attenuates the view-specific interferences. In Section \ref{sec:Audio-Visual-Fusion},
we describe the construction of a unified low dimensional representation
of the data through the eigenvalue decomposition of the matrix $\mathbf{M}$
similarly to obtaining a low dimensional representation of the data
in a single view using principal component analysis.

\section{Graph Theory Interpretation For Kernel Bandwidth Selection\label{sec:Kernel-Bandwidth-Selection}}

Recall that the affinity kernel $\mathbf{K}_{v}$ in \eqref{eq:K_v}
defines a graph on $\left\{ \mathbf{v}_{n}\right\} _{n=1}^{N}$ such
that each data point $\mathbf{v}_{n}$ is a vertex and $K_{v}(m,n)=\exp\left(-\frac{||\mathbf{v}_{n}-\mathbf{v}_{m}||^{2}}{\epsilon_{v}}\right)$
is the weight of the edge between vertex $n$ and vertex $m$. The
kernel bandwidth $\epsilon_{v}$ controls the connectivity of the
graph. When $||\mathbf{v}_{n}-\mathbf{v}_{m}||^{2}<\epsilon_{v}$,
high similarities are obtained between data points $n$ and $m$,
and they are considered connected; when $||\mathbf{v}_{n}-\mathbf{v}_{m}||^{2}\gg\epsilon_{v}$
the similarity between the points is negligible and we assume no edge
between the points. In order to capture the geometric structure of
the data, common practice is to set the kernel bandwidth such that
each data point is connected to at least one other point, i.e.:
\begin{equation}
\epsilon_{v}>\max_{m}\left[\min_{n}\left(||\mathbf{v}_{n}-\mathbf{v}_{m}||^{2}\right)\right].\label{eq:sigma min}
\end{equation}
This choice is a necessary condition for the graph defined on the
dataset to be connected such that there exists a path between every
pair of points. In turn, a connected graph is a necessary condition
for the eigenvectors of the affinity kernel to form a discrete orthogonal
basis. This property is typically used for the construction of low
dimensional representations \cite{coifman2006diffusion,von2007tutorial}.
Yet, the kernel bandwidth should be sufficiently small to prevent
the association of data points with different content. In \cite{keller2010audio},
the authors proposed choosing the value of the kernel bandwidth by:
\begin{equation}
\epsilon_{v}=C\cdot\max_{m}\left[\min_{n}\left(||\mathbf{v}_{n}-\mathbf{v}_{m}||^{2}\right)\right],\label{eq: sigma}
\end{equation}
where $C$ is a parameter which is set in the range of $2\div3$ to
guarantee that the graph is connected in the single view case such
that typically each point is connected to several other points. In
this study, we focus on the selection of the kernel bandwidth according
to \eqref{eq: sigma}, yet other existing methods for the kernel bandwidth
selection, e.g., those presented in \cite{zelnik2004self,von2007tutorial,coifman2008graph},
also rely on similar graph connectivity. 

In the multiple view case, the kernel bandwidth in each view is typically
set in the literature as if the data is captured only in a single
view and also require graph connectivity for each view, e.g, in \cite{de2010multi,wang2012unsupervised,lindenbaum2015multiview,lederman2015learning}.
In contrast, we show that when the data is measured in multiple views,
the graph of each single view does not necessarily have to be connected. 

To demonstrate this idea, we consider a multiple view graph, which
is defined by the Markov matrix $\mathbf{M}$ in \eqref{eq:M}. The
vertices of the graph are pairs of data points $\left\{ \left(\mathbf{v}_{n},\mathbf{w}_{n}\right)\right\} _{n=1}^{N}$,
and the matrix $\mathbf{M}$ defines a Markov chain on this graph
such that the $(n,m)$th entry of $\mathbf{M}$ is the probability
of a transition from vertex $n$ to vertex $m$. For simplicity, we
relate to (say) vertex $n$ as to point $n$ even though it is related
to the pair of points $\left(\mathbf{v}_{n},\mathbf{w}_{n}\right)$.
The matrix $\mathbf{M}$ aggregates the relations between the data
points based on the two views; there exists an edge between point
$n$ and point $m$ in the multiple view graph if the transition probability
between them, given by $\mathbf{M}(n,m)$, is non-zero. To capture
the geometric structure of the data, the necessary condition that
each point is connected to at least one other point applies to the
multiple view graph and not to the graphs of the single views. Namely,
the single view graphs can be disconnected while each point in the
multiple view graph is connected as demonstrated by Proposition \ref{prop:connectivity}. 
\begin{prop}
\label{prop:connectivity}$\forall n,\,\exists m\neq n$ such that
$M(n,m)\neq0$ iff $\forall n,\,\exists m\neq n$ such that $M_{v}(n,m)\neq0$
or $M_{w}(n,m)\neq0$. 
\end{prop}
Proposition \ref{prop:connectivity} implies that each point in the
multiple view graph is connected if it is connected at least in one
of the views. 
\begin{IEEEproof}
If point $n$ is disconnected in the first view, the $n$th row of
the affinity kernel of the first view $\mathbf{K}_{v}$ is given by:
\[
\begin{array}{c}
\left(K_{v}(n,1),K_{v}(n,2),...,K_{v}(n,n),...,K_{v}(n,N)\right)=\,\,\,\,\,\,\,\,\,\,\,\,\,\,\,\,\,\,\,\,\,\,\,\,\,\,\,\,\,\,\,\,\,\,\,\,\\
\,\,\,\,\,\,\,\,\,\,\,\,\,\,\,\,\,\,\,\,\,\,\,\,\,\,\,\,\,\,\,\,\,\,\,\,\,\,\,\,\,\,\,\,\,\,\,\,\,\,\,\,\,\,\,\,\,\,\,\,\,\,\,\,\,\,\,\,\,\,\,\,\,\,\,\,\,\,\,\,\,\,\,\,\,\,\,\,\,\,\,\,\,\left(0,0,...,1,...0\right).
\end{array}
\]
Consequently, the $n$th row of the corresponding row stochastic Markov
matrix $\mathbf{M}_{v}$ is given by:
\[
\begin{array}{c}
\left(M_{v}(n,1),M_{v}(n,2),...,M_{v}(n,n),...,M_{v}(n,N)\right)=\,\,\,\,\,\,\,\,\,\,\,\,\,\,\,\,\,\,\,\,\,\,\,\,\,\,\,\,\,\,\,\,\,\,\,\,\\
\,\,\,\,\,\,\,\,\,\,\,\,\,\,\,\,\,\,\,\,\,\,\,\,\,\,\,\,\,\,\,\,\,\,\,\,\,\,\,\,\,\,\,\,\,\,\,\,\,\,\,\,\,\,\,\,\,\,\,\,\,\,\,\,\,\,\,\,\,\,\,\,\,\,\,\,\,\,\,\,\,\,\,\,\,\,\,\,\,\,\,\,\,\left(0,0,...,1,...0\right).
\end{array}
\]
According to \eqref{eq:M} and by the rule of matrix product, the
$n$th row of $\mathbf{M}$ is given by:
\[
\begin{array}{c}
\left(0,0,...,1,...0\right)\cdot\mathbf{M}_{w}=\,\,\,\,\,\,\,\,\,\,\,\,\,\,\,\,\,\,\,\,\,\,\,\,\,\,\,\,\,\,\,\,\,\,\,\,\,\,\,\,\,\,\,\,\,\,\,\,\,\,\,\,\,\,\,\,\,\,\,\,\,\,\,\,\,\,\,\,\,\,\,\,\,\,\,\,\,\,\,\,\,\,\,\,\,\,\,\,\,\,\,\,\,\,\,\,\,\,\,\,\,\,\,\,\,\,\,\\
\left(M_{w}(n,1),M_{w}(n,2),...,M_{w}(n,n),...,M_{w}(n,N)\right).
\end{array}
\]
Therefore, $M(n,m)\neq0$ iff $M_{w}(n,m)\neq0$. If point $n$ is
connected to (say) point $m$ in the first view, i.e., $M_{v}(n,m)\neq0$,
by the matrix product rule, the $n$th row in $\mathbf{M}$ is given
by a linear combination of row $n$ and row $m$ in $\mathbf{M}_{w}$;
since $M_{w}(m,m)\neq0$ (each point is connected to itself), we have
that $M(n,m)\neq0$.
\end{IEEEproof}
Namely, the necessary condition to learn the geometry of the data
from two views is that each point is connected at least in one of
the views. Therefore, the kernel bandwidths of each view may be set
to small values without satisfying the requirement that the graphs
of the single views are connected. We will show in Section  \ref{sec:Simulation-Results}
that assigning small values to the kernel bandwidth increases the
robustness of the representation obtained using the multiple view
affinity kernel to interferences.

The remainder of this section revolves around the selection of the
kernel bandwidth. By using a simplifying statistical model for the
graph connectivity, we associate the selection of the kernel bandwidth
in the single view case with the average number of connections, which
we denote by $\delta$. Then we show that in the multiple view case
a proper kernel bandwidth is obtained by reducing the number of connections
up to a root factor, i.e., $\sqrt{\delta}$. Based on this result,
we present an algorithm for the selection of the kernel bandwidths.

Let $\mathds{1}_{v}\left(n,m\right)$ be an indicator which equals
one if point $n$ and point $m$ are connected in the first view and
zero otherwise. For simplicity, we assume that each pair of data points
is connected with probability $p_{v}$ independently from all other
data-points in the view. Namely, $\left\{ \mathds{1}_{v}\left(n,m\right)\right\} _{n,m}$
are independent and identically distributed (iid) random variables
such that: 
\begin{equation}
\mathds{1}_{v}\left(n,m\right)=\left\{ \begin{array}{cc}
1, & \text{w.p.}\,p_{v}\\
0, & \text{otherwise}
\end{array}\right\} .\label{eq:1_v}
\end{equation}
In addition, we assume that the connectivity between data points in
a certain view is independent from the connectivity in the other views.
We note that these two assumptions do not usually hold in practice.
For example, two points being connected to a third point implies that
the two points are close to each other, and as a result, they are
connected with high probability. In addition, since the data from
the different views are measurements of the same phenomenon, high
correlation is expected across the views. Yet, we justify these assumptions
by considering data contaminated with interferences, and assuming
that the interferences reduce these correlations.

Based on this statistical model, the number of connections of a certain
point to the other $N-1$ points in the graph of the first view is
given by a binomial distribution, denoted by $B_{v}$:
\[
B_{v}\left(N-1,p_{v}\right).
\]
The parameter $p_{v}$ is directly related to the kernel bandwidth
$\sigma_{v}$ in \eqref{eq:K_v}; the larger the kernel bandwidth,
the higher the probability that two points are connected. We assume
that the kernel bandwidth, and therefore $p_{v}$, are chosen such
that each point is connected on average to $S_{v}$ points, i.e.,
$p_{v}\approx\frac{S_{v}}{N-1}$, where $p_{v}\cdot(N-1)$ is the
mean value of the binomial distribution. Based on this model, the
probability that a certain point is disconnected is denoted by $q_{v}$
and is given by:
\[
q_{v}=\left(1-p_{v}\right)^{N-1}.
\]
For large values of $N$, we approximate $q_{v}$ by: 
\begin{equation}
q_{v}\thickapprox\left(1-\frac{S_{v}}{N-1}\right)^{N-1}\thickapprox e^{-S_{v}}.\label{eq:e^-sv}
\end{equation}
We note that we assumed in \eqref{eq:e^-sv} that the average number
of connections $S_{v}$ does not depend on the number of data points
$N$. In fact, some studies, e.g., the one presented in \cite{zelnik2004self},
suggest setting a constant number of connections to each point regardless
to the size of the dataset. 

In the single view case, the kernel bandwidth $\epsilon_{v}$ is chosen
such that the graph is connected. Under this statistical model, it
is equivalent to setting $S_{v}$ such that the probability $q_{v}$
in \eqref{eq:e^-sv} approaches zero. Namely, setting the kernel bandwidth
is equivalent to setting the average number of connections $S_{v}$
to a certain value $\delta$ such that $e^{-\delta}$ approaches zero. 

We proceed by considering the multiple view case, in which a similar
statistical model is considered for the second view as well. Let $p_{w,}\,S_{w}$
and $q_{w}$ be the equivalents of $p_{v,}\,S_{v}$ and $q_{v}$ in
the second view, respectively. We recall that a pair of points, point
$n$ and point $m$, is connected in the multiple view graph if the
$\left(n,m\right)$th entry of $\mathbf{M}$ in \eqref{eq:M} is non-zero.
The $(n,m)$th entry is explicitly written as: 
\[
M\left(n,m\right)={\displaystyle \sum_{l}M_{v}\left(n,l\right)M_{w}\left(l,m\right).}
\]
which implies that point $n$ and point $m$ are connected in the
multiple view graph if there exists a third point $l$ such that points
$n$ and $l$ are connected in the first view and points $l$ and
$m$ are connected in the second view. Since the probability that
the pair is connected via a third point $l$ is $p_{v}p_{w}$, there
will be on average $\left(N-2\right)p_{v}p_{w}$ such points, where
for simplicity we assumed that $l\neq n\neq m$. The term $\left(N-2\right)p_{v}p_{w}$
may be rewritten as: 
\[
\left(N-2\right)p_{v}p_{w}=\left(N-2\right)\frac{S_{v}}{N-1}\frac{S_{W}}{N-1}\approx\frac{S_{v}S_{W}}{N-1},
\]
where we neglect the term $\frac{N-2}{N-1}$, which approximately
equals one for large values of $N$. Typically, the average number
of points connected to a certain point, $S_{v}$ in the first view
or $S_{w}$ in the second view, is significantly smaller than $N$.
Therefore, we assume that $\frac{S_{v}S_{W}}{N-1}<1$, and we view
this term as the probability that point $n$ and point $m$ are connected
in the multiple view graph. The number of connections of each point
in the multiple view graph is therefore given by the following binomial
distribution:
\begin{equation}
B\left(N-1,\frac{S_{v}S_{w}}{N-1}_{v}\right),\label{eq:B}
\end{equation}
and, specifically, each point in the multiple view graph is connected
on average to $S_{v}S_{w}$ other points. Based on the binomial distribution
and similarly to \eqref{eq:e^-sv}, the probability that a point is
disconnected in the multiple view graph, which we denote by $q$,
is approximated by:
\[
q\thickapprox\left(1-\frac{S_{v}S_{w}}{N-1}\right)^{N-1}\thickapprox e^{-S_{v}S_{W}}.
\]
We interpret this result similarly to the result obtained for the
single view case; to meet the condition that each point in the graph
is connected, the probability $q$ has to approach zero, i.e., $q=e^{-\delta}$
as in the single view case, such that $S_{v}S_{w}=\delta$. Assuming
for simplicity that $S_{v}=S_{w}$, the average number of connections
in each view should be set to $S_{v}=S_{w}=\sqrt{\delta}$. In summary,
in the multiple view case, we may reduce the average number of connections
of each point by a root factor, and thus, significantly reduce the
size of the kernel bandwidth, while meeting the requirement on the
connectivity of the multiple view graph. We will show in Section \ref{sec:Simulation-Results}
that this choice of the kernel bandwidth improves the representation
obtained by the multiple view kernel method.

Next, we describe an algorithm for the selection of the kernel bandwidths.
For simplicity, we consider the selection of the kernel bandwidth
of the first view; the selection of the kernel bandwidth of the second
view is equivalent. We start by estimating $\delta$, i.e., the average
number of connections, $S_{v}=\left(N-1\right)p_{v}$, when the kernel
bandwidth is selected according to \eqref{eq: sigma} as if the data
is captured only in a single view. Recalling that $p_{v}$ is the
probability that two arbitrary points are connected, we estimate it
by:
\begin{equation}
\hat{p}_{v}=\frac{1}{N(N-1)}{\displaystyle \sum_{m}\sum_{n\neq m}K_{v}\left(n,m\right),}\label{eq:p^}
\end{equation}
where $\hat{p}_{v}$ is the estimate of $p_{v}$, and $K_{v}(n,m)$
is the $\left(n,m\right)$th entry of the affinity kernel $\mathbf{K}$
in \eqref{eq:K_v}. According to \eqref{eq:K_v}, $K_{v}\left(n,m\right)$
is in the range of $0\div1$ and a high value of $K_{v}\left(n,m\right)$
indicates that points $n$ and $m$ are connected. By selecting the
kernel bandwidth according to \eqref{eq: sigma} as if the data captured
in a single view, the estimate of $\delta$, denoted by $\hat{\delta}$,
is given by:
\begin{equation}
\hat{\delta}=\left(N-1\right)\hat{p}_{v}=\frac{1}{N}{\displaystyle \sum_{m}\sum_{n\neq m}K_{v}\left(n,m\right).}\label{eq:delta^}
\end{equation}
We denote the new bandwidth of the affinity kernel by $\epsilon_{v}^{\text{AD}}$,
where $\text{AD}$ is alternating diffusion, and we select it such
that the estimated average number of connections, which we denote
by $\delta^{\text{AD}}$, is reduced to $\sqrt{\hat{\delta}}$. We
propose selecting $\epsilon_{v}^{\text{AD}}$ similarly to \eqref{eq: sigma}
by:
\begin{equation}
\epsilon_{v}^{\text{AD}}=C^{\text{AD}}\cdot\max_{m}\left[\min_{n}\left(||\mathbf{v}_{n}-\mathbf{v}_{m}||^{2}\right)\right],\label{eq:sigma_tilde}
\end{equation}
where $C^{\text{AD}}$ is a parameter in the range of $0\div C$.
The selection of a proper kernel bandwidth is reduced to the selection
of the parameter $C^{\text{AD}}$ decreasing the average number of
connections by a root factor. We recall that to estimate $\hat{\delta}$
in \eqref{eq:delta^}, the parameter $C$ in \eqref{eq: sigma} is
chosen in the range of $2\div3$ (as if the data is captured in a
single view), and propose searching the parameter $C^{\text{AD}}$
within a discrete set whose elements lie on a linear grid. Let $\mathcal{C}=\left\{ C_{k}\right\} _{k=1}^{\left|\mathcal{C}\right|}$
be a discrete set of size $\left|\mathcal{C}\right|$, where $C_{k},\,k=1,2,...,\left|\mathcal{C}\right|$
are given by $C_{k}=\frac{k}{\left|\mathcal{C}\right|}C$. We propose
applying a binary search within the set such that the proposed kernel
bandwidth, $C^{\text{AD}}$, is given by the element $C_{k}$ for
which the average number of connections $\delta^{\text{AD}}$ is the
closest to $\sqrt{\hat{\delta}}$. We summarize the proposed algorithm
in Algorithm \ref{alg:Kernel-bandwidth-selection}, and note that
the kernel bandwidth in the second view is selected similarly.

\begin{algorithm}
\protect\caption{Kernel bandwidth selection\label{alg:Kernel-bandwidth-selection}}

\begin{algorithmic}[1] 

\State{Calculate $\epsilon_v$ in \eqref{eq: sigma} by setting $C=2\div3$ as if the data is captured in a single view}
\State{Calculate $\mathbf{K}_v$ in \eqref{eq:K_v}}
\State{Estimate $\hat{\delta}$ in \eqref{eq:delta^}}
\State{Define: $\mathcal{C}=\left\{ C_{k}\right\} _{k=1}^{\left|\mathcal{C}\right|}$, where $C_k=\frac{k}{\left|\mathcal{C}\right|}C$}

\While{$|\mathcal{C}|\neq{1}$} 
	\State{$C^{\text{AD}}=C_{|\mathcal{C}|/2}$}
	\State{Estimate $\delta^{\text{AD}}$ similarly to \eqref{eq:delta^} by recalculating $\mathbf{K}_v$ with the parameter $C^{\text{AD}}$}
	\If{$\delta^{\text{AD}}>\sqrt{\hat\delta}$}
		\State{$\mathcal{C}=\left\{ C_{k}\right\} _{k=1}^{\left|\mathcal{C}\right|/2}$}
	\Else
				\State{$\mathcal{C}=\left\{ C_{k}\right\} _{k={\left|\mathcal{C}\right|/2+1}}^{\left|\mathcal{C}\right|}$}
	\EndIf
\EndWhile 
\State{Using the obtained $C^{\text{AD}}$, calculate the new kernel bandwidth $\epsilon_v^{\text{AD}}$ in \eqref{eq:sigma_tilde}}
\end{algorithmic}
\end{algorithm}

\section{Audio Visual Fusion With Application To Voice Activity Detection\label{sec:Audio-Visual-Fusion}}

We consider speech measured by a single microphone and by a video
camera pointed to the face of the speaker. The audio-visual signal
is processed in frames, and we consider a sequence of $N$ frames,
which are aligned in the two views (audio and video). While speech
is measured in the two views, the interruptions are view specific.
The audio signal consists of, in addition to speech, background noise
and transients, which are short-term interferences, e.g. keyboard
taps and office noise; the video signal contains mouth movements during
non-speech intervals, which are considered as interferences since
they appear similar to speech. Our goal is obtaining a representation
of speech, which is robust to noise and interferences. In order to
accomplish this goal, we apply alternating diffusion maps, where the
kernel bandwidth is determined according to Algorithm \ref{alg:Kernel-bandwidth-selection}.

The alternating diffusion maps method is applied in a domain of features,
which are designed to reduce the effect of the interferences \cite{dov2015audio}.
The audio signal is regarded as the first view, and it is represented
by features based on \ac{MFCC}, which are commonly used for speech
representation \cite{logan2000mel}. Specifically, the $n$th data
point in the first view, $\mathbf{v}_{n}\in\mathbb{R}^{L_{v}}$ in
\eqref{eq:dataset}, is the feature vector of the $n$th frame, and
is given by the concatenation of the MFCCs of frames $n-1,n,$ and
$n+1$. Namely, $L_{v}$ is the total number of the coefficients in
three consecutive frames. The use of consecutive frames reduces the
effect of transients since speech is assumed more consistent over
time compared to transients, which are rapidly varying. The data obtained
in the second view, i.e., the video signal, is represented by motion
vectors \cite{bruhn2005lucas} such that the production of speech
is assumed associated to high levels of mouth movement. The features
representation of the $n$th frame of the video signal, $\mathbf{w}_{n}\in\mathbb{R}^{L_{w}}$,
is given by concatenating the absolute values of the motion vectors
in frames $n-1,n$ and $n+1$. Similarly to the audio signal, the
use of consecutive frames for representation reduces the effect of
short-term mouth movements during non-speech intervals. For more details
on the construction of the features, we refer the reader to \cite{dov2015audio}. 

The representation using the specifically designed features is only
partly robust to the interferences. For example, video features of
a non-speech frame may be wrongly similar to the features of a speech
frame if the former contains large movements of the mouth. To further
improve the robustness of the representation to noise, the two views
are fused using the alternating diffusion maps method with the improved
affinity kernel. Specifically, we construct the affinity kernels of
the two views, $\mathbf{M}_{v}$ and $\mathbf{M}_{w}$, according
to \eqref{eq:K_v} and \eqref{eq:M_v} using the features $\left\{ \mathbf{v}_{n}\right\} _{n=1}^{N}$and
$\left\{ \mathbf{w}_{n}\right\} _{n=1}^{N}$, respectively, and fuse
the views by $\mathbf{M}=\mathbf{M}_{v}\cdot\mathbf{M}_{w}$. Then,
we construct an eigenvalue decomposition of $\mathbf{M}$ such that
the eigenvectors aggregate the connections between the data points
within each view and between the views into a global representation
of the data. Since the matrix $\mathbf{M}$ is row stochastic, the
eigenvalue with the largest absolute value is $1$ and it corresponds
to an all ones eigenvector \cite{coifman2006diffusion}. This eigenvector
is neglected since it does not contain information. We note that the
eigenvectors of $\mathbf{M}$ are not guaranteed to be real valued
as it is guaranteed for the single view matrices $\mathbf{M}_{v}$
and $\mathbf{M}_{w}$ since the latter are similar to symmetric matrices.
Therefore, one solution is constructing an eigenvalue decomposition
of the symmetric matrix $\mathbf{M}\mathbf{M}^{T}$. A different approach
is obtaining a representation by the singular value decomposition
of $\mathbf{M}$ as described in \cite{Michaeli2016nonparametric}.
Yet, our experiments have shown that the eigenvectors corresponding
to the several largest eigenvalues of $\mathbf{M}$ are indeed real
and that the three approaches perform similarly.

We demonstrate the use of the representation obtained by alternating
diffusion maps for the problem of voice activity detection. Let $\mathbb{\mathcal{H}}_{0},\mathbb{\mathcal{H}}_{1}$
be hypotheses of speech absence and speech presence, respectively,
and let $\mathds{1}_{n}$ denote a speech indicator at the $n$th
frame, given by:
\[
\mathds{1}_{n}=\left\{ \begin{array}{ccc}
1 & ; & \mathcal{H}_{1}\\
0 & ; & \mathcal{H}_{0}
\end{array}\right\} .
\]
Given a sequence of $N$ frames, the goal is to estimate the speech
indicator, i.e., to separate the sequence of frames to speech and
non-speech clusters. We found in our experiments that the obtained
representation of the audio-visual signal, and specifically, its first
coordinate, i.e., the leading (non-trivial) eigenvector of the matrix
$\mathbf{M}$ in \eqref{eq:M}, which we denote by $\boldsymbol{\nu}_{1}\in\mathbb{R}^{N}$,
successfully separates between speech and non-speech frames. Therefore,
we take a similar approach to \cite{dov2015anisotropic} and estimate
the speech indicator by comparing the leading eigenvector to a threshold
$\tau$:
\begin{equation}
\hat{\mathds{1}}{}_{n}=\left\{ \begin{array}{ccc}
1 & ; & \nu_{1}(n)>\tau\\
0 & ; & \text{otherwise}
\end{array}\right\} ,\label{eq:est_indicator}
\end{equation}
where $\nu_{1}(n)$ in the $n$th entry of the eigenvector $\boldsymbol{\nu}_{1}$,
and the threshold may be chosen according to a specific application.
We note that the leading eigenvector $\boldsymbol{\nu}_{1}$ is widely
used in the literature for clustering and it was shown in \cite{shi2000normalized}
that it solves the well-known normalized cut problem. In contrast
to previous works, in this study the leading eigenvector is obtained
from the multiple view Markov matrix such that it clusters the data
according to the two views. Similarly to \cite{dov2015anisotropic},
the leading eigenvector  is used as a continuous measure of voice
activity rather than for binary clustering. The proposed voice activity
detection algorithm is summarized in Algorithm \ref{alg:vad}. Before
proceeding to the experimental results, we note that the proposed
representation and hence the speech indicator are obtained in a batch
manner assuming that $N$ consecutive frames are available in advance.
Yet, as described in \cite{dov2015audio}, a training set may be used
to construct the representation, and then it can be extended to new
incoming frames, e.g., using the Nyström method, in an online manner
\cite{fowlkes2004spectral}. 

\begin{algorithm}
\protect\caption{Voice activity detection\label{alg:vad}}

\begin{algorithmic}[1] 

\State{Calculate the features of the audio-visual signal $\left(\mathbf{v}_{1},\mathbf{w}_{1}\right),\left(\mathbf{v}_{2},\mathbf{w}_{2}\right),...,\left(\mathbf{v}_{N},\mathbf{w}_{N}\right)$}
\State{Calculate $\mathbf{K}_v$ and $\mathbf{K}_w$ according to \eqref{eq:K_v} and Algorithm \ref{alg:Kernel-bandwidth-selection}}
\State{Calculate $\mathbf{M}_v$ and $\mathbf{M}_w$ according to \eqref{eq:M_v}}
\State{Fuse the views by calculating $\mathbf{M}$ in \eqref{eq:M}}
\State{Obtain the leading eigenvector  $\boldsymbol{\nu}_1$}
\For{$n=1:N$} 
	\If{$\nu_1(n)>\tau$}
		\State{$\hat{\mathds{1}}{}_{n}=1$}
	\Else
		\State{$\hat{\mathds{1}}{}_{n}=0$}
	\EndIf
\EndFor
\end{algorithmic}
\end{algorithm}

\section{Simulation Results\label{sec:Simulation-Results}}

We use a dataset that we recently presented in \cite{dov2015audio}.
The signals are recorded using a microphone and a frontal video camera
of a smartphone pointed to the face of the speaker. The video signal
is processed in $25$ fps and it comprises the region of the mouth
of the speaker automatically cropped out from the recorded video as
described in \cite{dov2015audio} and illustrated in Fig. \ref{fig:example}.
The audio signal is processed in $8$ kHz using time frames of $634$
samples with $50\%$ overlap, such that this setup aligns between
the audio and the video signals. The dataset comprises $11$ sequences
of different speakers, each of which is $60$ s long containing speech
and non-speech intervals. 

\begin{center}
\begin{figure}
\begin{centering}
\includegraphics[scale=0.6]{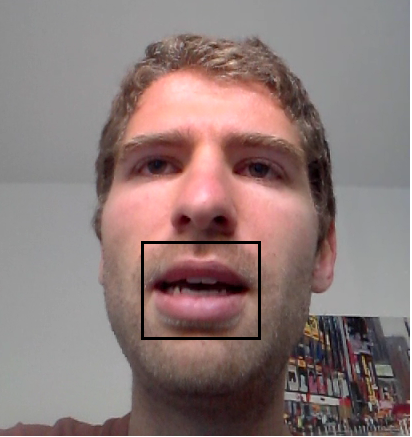}
\par\end{centering}

\centering{}\protect\caption{An example of a video frame and the cropped mouth region.\label{fig:example}}
\end{figure}

\par\end{center}

The signals are recorded in a quiet room and we synthetically add
different types of background noise and transients to the audio signal.
The transients are taken from an online free corpus \cite{freesound}
and they are normalized such that they have the same maximal amplitude
as the clean audio signal. Based on the clean audio, we mark the ground
truth in each frame, such that frames with energy level higher than
$1\%$ of the highest energy level in the sequence are marked as speech
frames. This setup of voice detection has a fine resolution of few
tens of milliseconds, and it is useful for application such as speech
recognition where single phonemes should be isolated \cite{rabiner1993fundamentals,ramirez2007voice}.

In the first experiment, we evaluate the proposed voice activity detection
algorithm, termed ``Alternating'' in the plots, by comparing it to
other versions of the algorithm based only on a single view (audio
or video). We term the versions of the algorithm based on the first
and the second view ``Audio'' and ``Video'', respectively, and the
corresponding speech indicators are estimated by comparing the leading
eigenvectors of the matrices $\mathbf{M}_{v}$ in \eqref{eq:M_v}
and $\mathbf{M}_{w}$ to a threshold, respectively. In addition, we
examine another two approaches for the fusion of the views using the
corresponding row stochastic matrices. In the first approach, the
fused matrix is given by the Hadamard product between the matrices:
$\mathbf{M}_{v}\circ\mathbf{M}_{w}$, where $\circ$ denotes point-wise
multiplication, and in the second approach the views are fused by
a simple sum: $\mathbf{M}_{v}+\mathbf{M}_{W}$. These approaches are
termed in the plots ``Hadamard'' and ``Sum'', respectively. We note
that both in the proposed algorithm and in the competing methods,
the speech indicator is estimated by the leading eigenvector obtained
by the eigenvalue decomposition with an arbitrary sign. To set the
sign of the eigenvector, one may for example consider the variability
of the video signal over time such that the lack of mouth movement
over several consecutive frames indicates absence of speech. In this
study, the sign of the eigenvector is assumed to be known for all
the methods. 

In addition to the different merging schemes, we compare the proposed
algorithm to the method presented in \cite{tamura2010robust} termed
``Tamura'' in the plots. The performances of the algorithms are
presented in Fig. \ref{fig:ROC} for different types of transients
in the form of ROC curves, i.e., plots of probability of detection
versus probability of false alarms. The larger the \ac{AUC} is, the
better the performance of the algorithms are, and the AUC of each
algorithm is presented in the legend box. It can be seen in the plots
that the algorithm based on the video signal provides relatively poor
performance compared to the other algorithms. This is mainly since
the ground truth is set to a fine resolution, and the video signal
is not sensitive enough. For example, video frames of a closed mouth
may be measured during both speech and non-speech intervals. We note
that in most of the previous studies, the video signal is used for
the detection of long speech intervals of several words, and it cannot
detect speech in fine resolutions. In addition, we note that we also
compared the proposed algorithm to the algorithm we recently presented
in \cite{dov2015audio}. In \cite{dov2015audio}, we proposed a separate
representation of each view and estimated voice activity separately
based on each representation; then, we fused the views by merging
the estimators. However, due to the challenging problem setting considered
in this study, for which the speech is detected at a fine resolution,
we found that incorporating the visual information as proposed in
\cite{dov2015audio} does not improve the detection scores. Hence
the simulation of \cite{dov2015audio} is not presented in the plots.

The audio signal in Fig. \ref{fig:ROC} also performs poorly due to
the presence of transients, which are not properly separated from
speech. The other fusion approaches, ``Hadamard'' and ``Sum'' slightly
benefit from the fusion of the sensors and provide performances comparable
to the performance obtained by audio signal. The proposed fusion of
the audio-visual signal provides improved performance and outperforms
all the other algorithms. 

\begin{center}
\begin{figure}
\begin{centering}
\subfloat[]{\protect\centering{}\protect\includegraphics[bb=30bp 180bp 550bp 590bp,clip,scale=0.48]{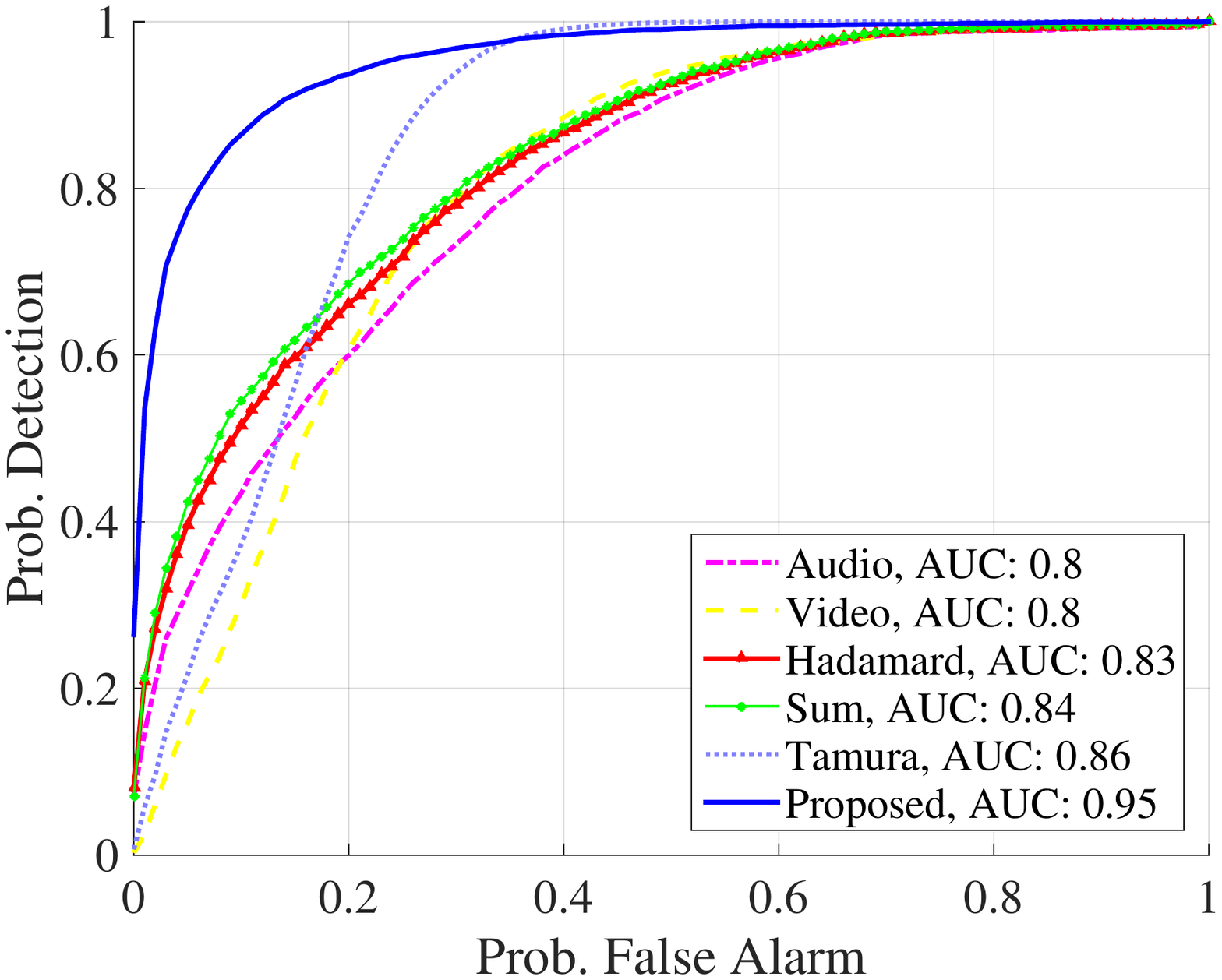}\protect}
\par\end{centering}

\begin{centering}
\subfloat[]{\protect\centering{}\protect\includegraphics[bb=30bp 180bp 550bp 590bp,clip,scale=0.48]{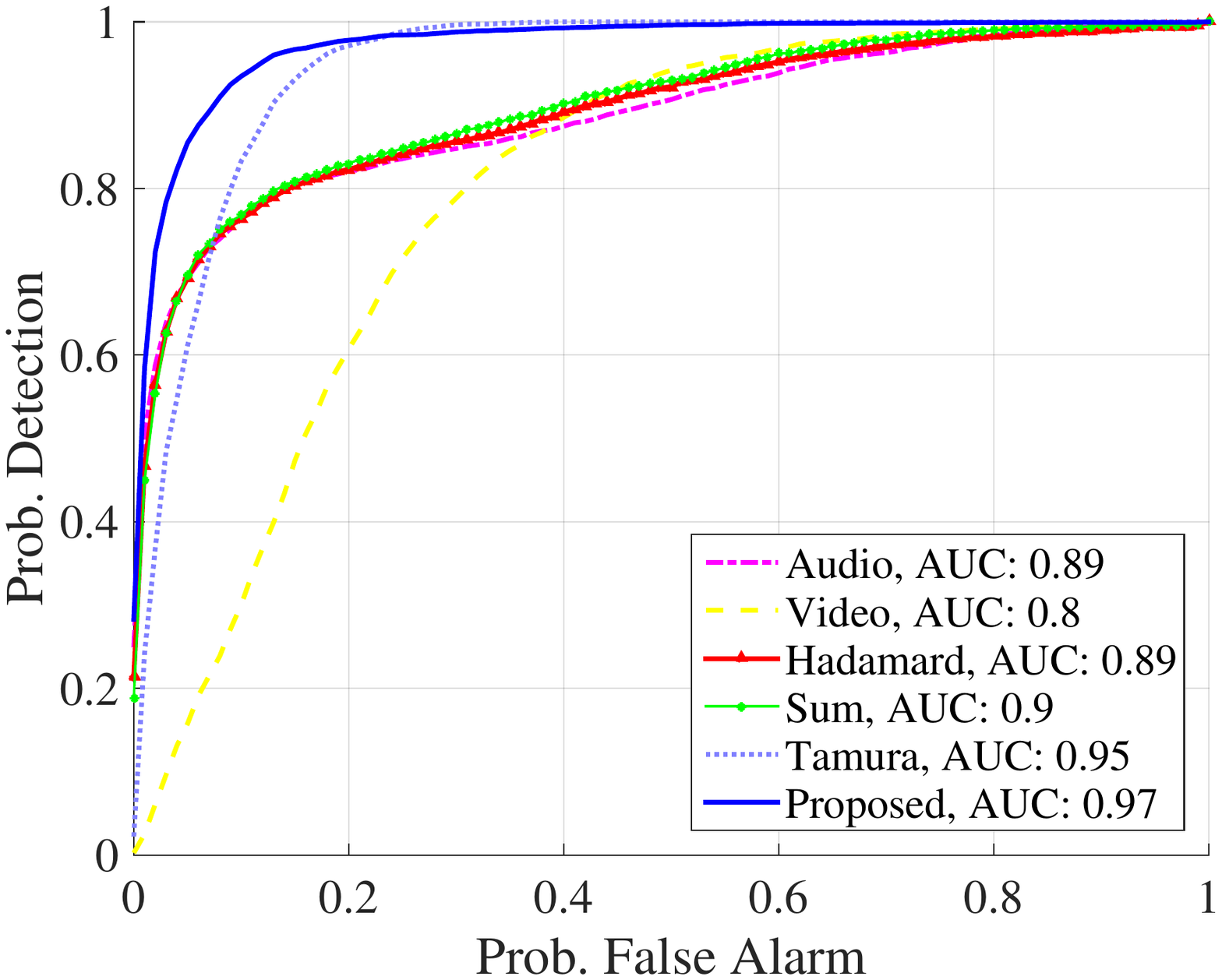}\protect}
\par\end{centering}

\begin{centering}
\subfloat[]{\protect\centering{}\protect\includegraphics[bb=30bp 180bp 550bp 590bp,clip,scale=0.48]{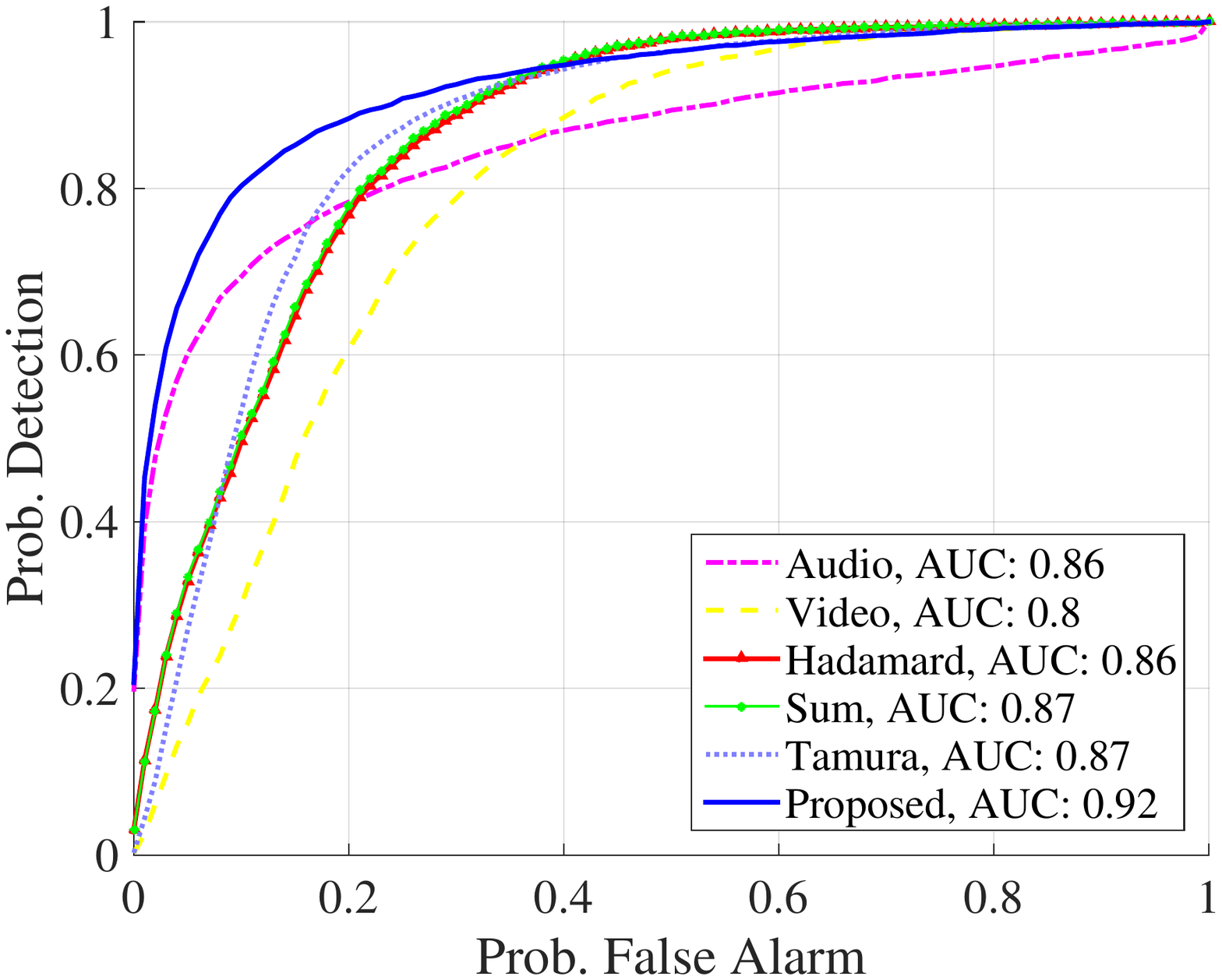}\protect}
\par\end{centering}

\protect\caption{Probability of the detection vs probability of false alarm. Transient
type: (a) hammering, (b) door-knocks, (c) microwave. \label{fig:ROC}}
\end{figure}

\par\end{center}

To further gain insight on the performance of the proposed algorithm
for voice activity detection, we present in Fig. \ref{fig:qualitative}
an example of speech detection in a sequence contaminated by hammering.
In this experiment, we set the threshold value in \eqref{eq:est_indicator}
to provide $90$ percent correct detection rate and compare the false
alarms resulting from the proposed algorithm to the false alarms resulting
from the algorithm presented in \cite{tamura2010robust}. As demonstrated
in Fig. \ref{fig:qualitative} (top), significantly less false alarms
are received by the proposed algorithm compared to the competing detector
such that the latter wrongly detects most of the transients as speech.

\begin{center}
\begin{figure}
\centering{}\includegraphics[bb=40bp 180bp 580bp 600bp,clip,scale=0.45]{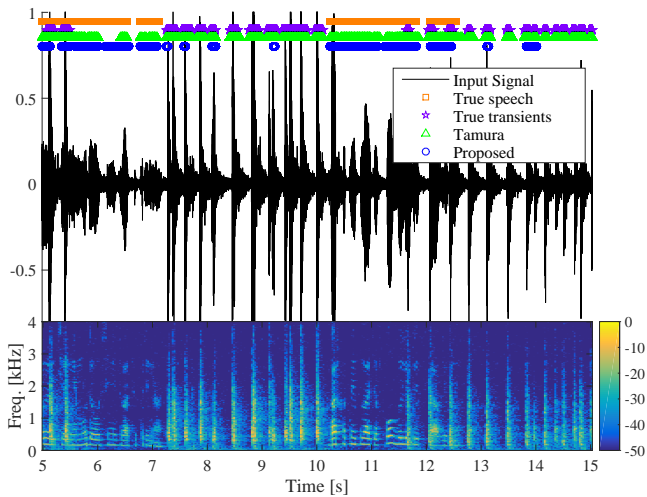}\protect\caption{Qualitative assessment of the proposed algorithm for voice activity
detection, with a hammering transient. (Top) Time domain, input signal-
black solid line, true speech- orange squares, true transients- purple
stars, ``Tamura'' with a threshold set for 90 percents correct detection
rate- green triangles, proposed algorithm with a threshold set for
90 percents correct detection rate- blue circles. (Bottom) Spectrogram
of the input signal.\label{fig:qualitative}}
\end{figure}

\par\end{center}

In Figs. \ref{fig:ROC}-\ref{fig:qualitative}, we set $C=2$ in \eqref{eq: sigma}
as if the data is obtained in a single view, and we proceed to evaluate
the performance of the proposed algorithm for different values of
the kernel bandwidth in the next experiment. In Fig. \ref{fig:AUC-vs-C}
we present plots of the AUC of the proposed voice activity detector
versus the parameter $C$ in \eqref{eq: sigma} for different types
of noise and interferences. We recall that the parameter $C$ represents
the kernel bandwidth such that in the single view case, connected
graphs typically correspond to $C$ values in the range $2\div3$,
and disconnected graphs correspond to $C$ values less than $1$.
The red solid line in Fig. \ref{fig:AUC-vs-C} is obtained by changing
only the parameter $C$ related to the audio signal while keeping
the parameter related to the video signal fixed (with a constant value
$C=2$). The blue dot in the plots is $C^{\text{AD}}$, i.e., the
proposed kernel bandwidth obtained by algorithm \ref{alg:Kernel-bandwidth-selection}.
We empirically found sufficient searching $C^{\text{AD}}$ over a
grid with a step of $\frac{C}{\left|\mathcal{C}\right|}=0.05$ such
that sweeping the parameter $C^{\text{AD}}$ over values below this
scale has negligible effect on the estimated average number of connections
in the graph. It can be seen in the plots that by reducing the value
of the parameter $C$ the AUC is improved up to a peak obtained when
$C\approx0.5$. The peak value in the plots is the sweet spot in the
trade-off in the kernel bandwidth selection. On the one hand, small
values of the kernel bandwidth remove wrong connections in the graph
between speech and non-speech frames, resulting in a representation
in which these frames are better separated. On the other hand, too
small kernel bandwidth causes the multiple view graph to be disconnected.
Indeed, the significant degradation of the AUC for parameter values
below the peak may indicate that the multiple view graph is disconnected
such that the obtained audio-visual representation no longer captures
the geometric structure of speech. The fact that the peak is obtained
for parameter value below $1$ indicates that a better representation
of the audio-visual signal is obtained by setting the kernel bandwidth
such that the graph of the audio signal is disconnected. These plots
demonstrate the idea that the kernel bandwidth should be chosen as
the smallest possible keeping the graph of the multiple views connected.
In addition, Fig. \ref{fig:AUC-vs-C} demonstrate the performance
of Algorithm \ref{alg:Kernel-bandwidth-selection} for the selection
of the kernel bandwidth. The parameter $C$ obtained by the algorithm,
i.e., $C^{\text{AD}}$, successfully provides AUC close to the peak
value. The slight deviation of $C^{\text{AD}}$ from the peak may
be explained by the assumptions on the statistical model in Section
\ref{sec:Kernel-Bandwidth-Selection}, which may not hold in practice. 

We note that Algorithm \ref{alg:Kernel-bandwidth-selection} is applied
to the audio view since we expect to benefit from the algorithm only
when there are high levels of noise and interferences. The video signal
is considered relatively clean even though there exist some non-speech
mouth movements, which may be wrongly detected as speech. In the case
of clean signals, audio or video, there are significantly less wrong
connections in the graphs of the single views, and hence, reducing
the kernel bandwidths does not improve the obtained representation
as in the case where the signal is measured in the presence of noises
and interferences. 

\begin{center}
\begin{figure}
\begin{centering}
\subfloat[]{\protect\centering{}\protect\includegraphics[bb=30bp 180bp 550bp 590bp,clip,scale=0.48]{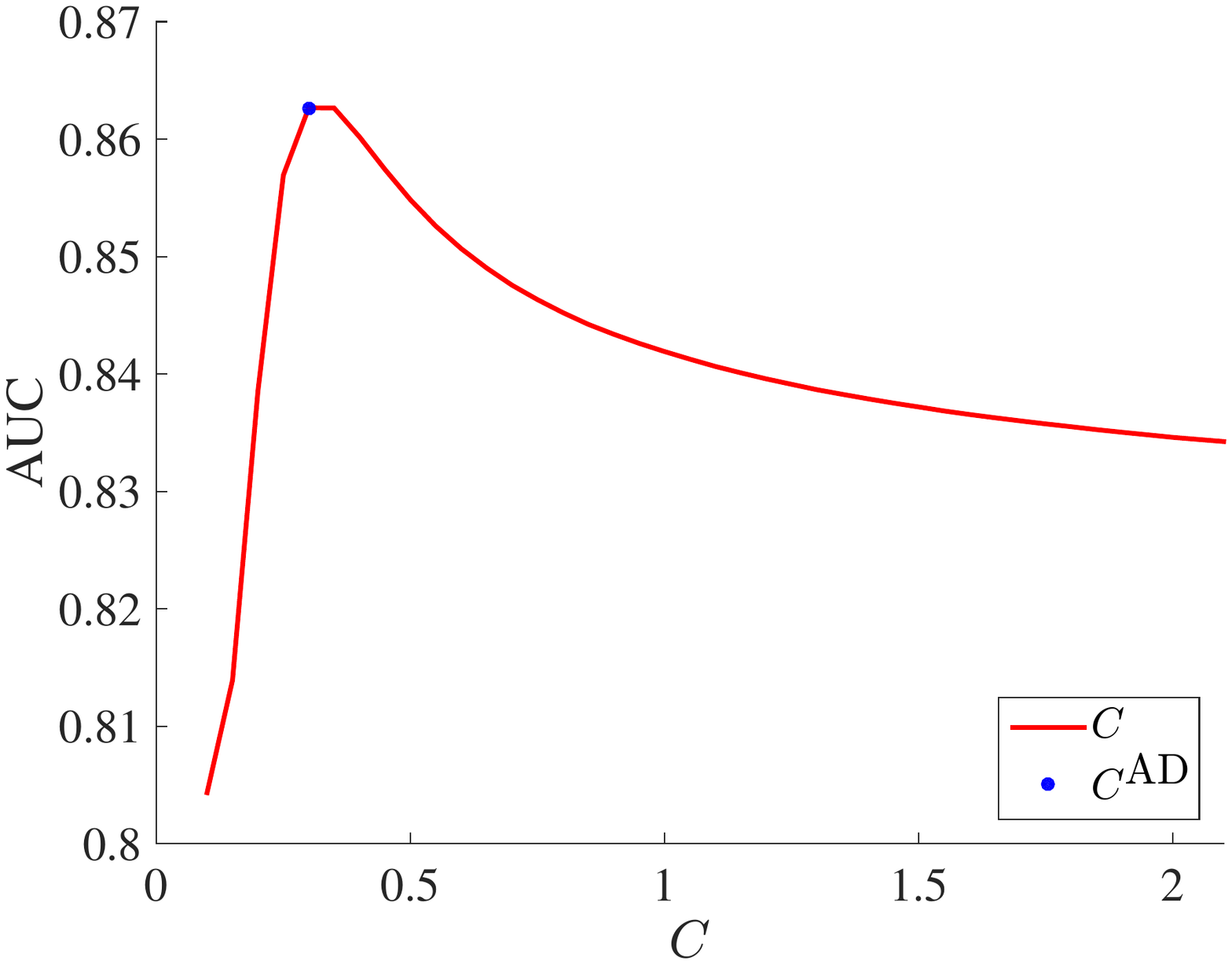}\protect}
\par\end{centering}

\begin{centering}
\subfloat[]{\protect\centering{}\protect\includegraphics[bb=30bp 180bp 550bp 590bp,clip,scale=0.48]{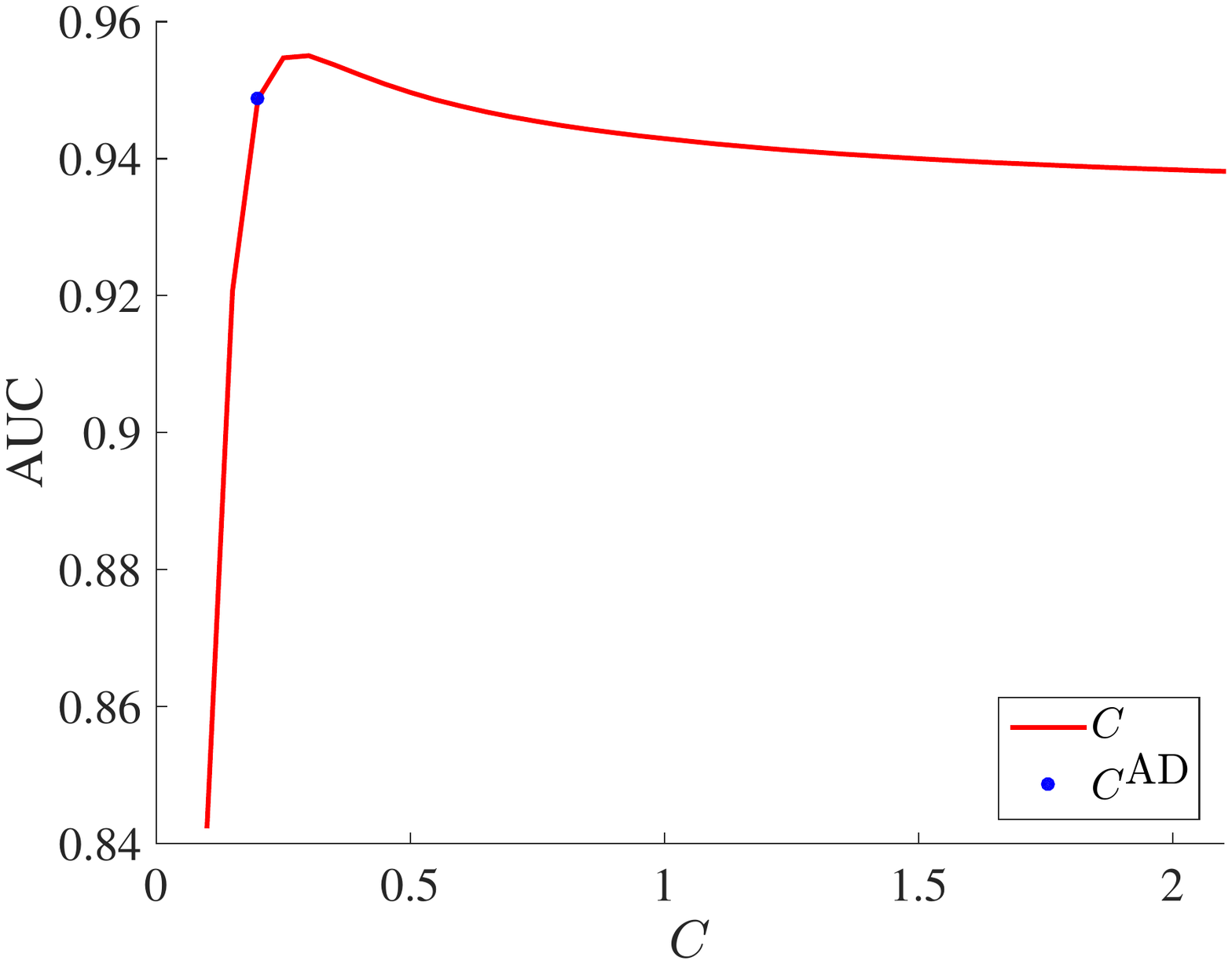}\protect}
\par\end{centering}

\protect\caption{AUC vs the parameter $C$ of the audio view. (a) background noise:
babble noise with $-5$ dB SNR, (b) transient type: keyboard-taps,
background noise: colored Gaussian noise with $-5$ dB SNR.\label{fig:AUC-vs-C}}
\end{figure}

\par\end{center}

\section{Conclusions}

We have addressed the problem of multiple view data fusion. We revisited
the alternating diffusion maps method and proposed a new interpretation
from a graph theory point of view, in which the affinity kernels of
the single and multiple views define graphs on the data. By introducing
a statistical model of the connectivity between data points on the
graphs, we showed that fusing the data by a product of the affinity
kernels increases the average number of connections in the multiple
view case. Accordingly, the kernel bandwidth, controlling the connectivity
between the data points, may be set significantly smaller than in
the single view case. Specifically, we showed that the proper kernel
bandwidth is the one reducing the average number of connections by
a root factor, and presented an algorithm for its selection. Using
the alternating diffusion maps method with the improved affinity kernel,
we have addressed the problem of audio-visual fusion. In particular,
we have considered the task of voice activity detection in the presence
of transients; we have shown that the representation obtained by alternating
diffusion maps allows for accurate speech detection using the first
coordinate, i.e., the leading eigenvector. Our simulation results
have demonstrated that the incorporation of visual data significantly
improves the detection scores both compared to detecting speech based
on only the audio data and compared to alternative merging schemes.
In addition, our simulation results have demonstrated that reducing
the kernel bandwidth below the values typically used in the single
view case improves the robustness of the fusion to transient interferences
and consequently the voice activity detection scores. Finally, we
have demonstrated that the proposed algorithm for the kernel bandwidth
selection allows for selecting near optimal values of the kernel bandwidth.

\bibliographystyle{../utils/IEEEbib}
\bibliography{papers}

\begin{thebibliography}{10}

\bibitem{roweis2000nonlinear}
S.~T. Roweis and L.~K. Saul,
\newblock ``Nonlinear dimensionality reduction by locally linear embedding,''
\newblock {\em Science}, vol. 290, no. 5500, pp. 2323--2326, 2000.

\bibitem{balasubramanian2002isomap}
M.~Balasubramanian, E.~L. Schwartz, Tenenbaum~J. B., de~Silva~V., and J.~C.
  Langford,
\newblock ``The isomap algorithm and topological stability,''
\newblock {\em Science}, vol. 295, no. 5552, pp. 7--7, 2002.

\bibitem{belkin2003laplacian}
M.l Belkin and P.~Niyogi,
\newblock ``Laplacian eigenmaps for dimensionality reduction and data
  representation,''
\newblock {\em Neural Computation}, vol. 15, no. 6, pp. 1373--1396, 2003.

\bibitem{donoho2003hessian}
D.~L. Donoho and C.~Grimes,
\newblock ``Hessian eigenmaps: Locally linear embedding techniques for
  high-dimensional data,''
\newblock {\em Proc. the National Academy of Sciences}, vol. 100, no. 10, pp.
  5591--5596, 2003.

\bibitem{coifman2006diffusion}
R.R. Coifman and S.~Lafon,
\newblock ``Diffusion maps,''
\newblock {\em Applied and Computational Harmonic Analysis}, vol. 21, no. 1,
  pp. 5--30, 2006.

\bibitem{mishne2013multiscale}
G.~Mishne and I.~Cohen,
\newblock ``Multiscale anomaly detection using diffusion maps,''
\newblock {\em IEEE Journal of Selected Topics in Signal Processing}, vol. 7,
  no. 1, pp. 111--123, 2013.

\bibitem{mishne2015graph}
G.~Mishne, R.~Talmon, and I.~Cohen,
\newblock ``Graph-based supervised automatic target detection,''
\newblock {\em IEEE Transactions on Geoscience and Remote Sensing}, vol. 53,
  no. 5, pp. 2738--2754, 2015.

\bibitem{talmon2012supervised}
R.~Talmon, I.~Cohen, S.~Gannot, and R.~R. Coifman,
\newblock ``Supervised graph-based processing for sequential transient
  interference suppression,''
\newblock {\em IEEE Transactions on Audio, Speech, and Language Processing},
  vol. 20, no. 9, pp. 2528--2538, 2012.

\bibitem{talmon2013single}
R.~Talmon, I.~Cohen, and S.~Gannot,
\newblock ``Single-channel transient interference suppression with diffusion
  maps,''
\newblock {\em IEEE Transactions on Audio, Speech, and Language Processing},
  vol. 21, no. 1, pp. 132--144, 2013.

\bibitem{zhou2007spectral}
D.~Zhou and C.~J.~C. Burges,
\newblock ``Spectral clustering and transductive learning with multiple
  views,''
\newblock in {\em Proceedings of the 24th international conference on Machine
  learning}. ACM, 2007, pp. 1159--1166.

\bibitem{blaschko2008correlational}
M.~B. Blaschko and C.~H. Lampert,
\newblock ``Correlational spectral clustering,''
\newblock in {\em IEEE Conference on Computer Vision and Pattern Recognition
  (CVPR), 2008.} IEEE, 2008, pp. 1--8.

\bibitem{de2010multi}
V.~R. De~Sa, P.~W. Gallagher, J.~M. Lewis, and V.~L. Malave,
\newblock ``Multi-view kernel construction,''
\newblock {\em Machine learning}, vol. 79, no. 1-2, pp. 47--71, 2010.

\bibitem{kumar2011co}
A.~Kumar, P.~Rai, and H.~Daume,
\newblock ``Co-regularized multi-view spectral clustering,''
\newblock in {\em Advances in Neural Information Processing Systems}, 2011, pp.
  1413--1421.

\bibitem{kumar2011co2}
A.~Kumar and H.~Daum{\'e},
\newblock ``A co-training approach for multi-view spectral clustering,''
\newblock in {\em Proceedings of the 28th International Conference on Machine
  Learning (ICML-11)}, 2011, pp. 393--400.

\bibitem{lin2011multiple}
Y.~Y. Lin, T.~L. Liu, and C.~S0 Fuh,
\newblock ``Multiple kernel learning for dimensionality reduction,''
\newblock {\em IEEE Transactions on Pattern Analysis and Machine Intelligence},
  vol. 33, no. 6, pp. 1147--1160, 2011.

\bibitem{wang2012unsupervised}
B.~Wang, J.~Jiang, W.~Wang, Z.~H. Zhou, and Z.~Tu,
\newblock ``Unsupervised metric fusion by cross diffusion,''
\newblock in {\em IEEE Conference on Computer Vision and Pattern Recognition
  (CVPR), 2012}. IEEE, 2012, pp. 2997--3004.

\bibitem{huang2012affinity}
H.~C. Huang, Y.~Y. Chuang, and C.~S. Chen,
\newblock ``Affinity aggregation for spectral clustering,''
\newblock in {\em IEEE Conference on Computer Vision and Pattern Recognition
  (CVPR), 2012}. IEEE, 2012, pp. 773--780.

\bibitem{boots2012two}
B.~Boots and G.~Gordon,
\newblock ``Two-manifold problems with applications to nonlinear system
  identification,''
\newblock {\em arXiv preprint arXiv:1206.4648}, 2012.

\bibitem{bronstein2013making}
M.~M. Bronstein, K.~Glashoff, and T.~A. Loring,
\newblock ``Making laplacians commute,''
\newblock {\em arXiv preprint arXiv:1307.6549}, 2013.

\bibitem{lindenbaum2015multiview}
Ofir Lindenbaum, Arie Yeredor, Moshe Salhov, and Amir Averbuch,
\newblock ``Multiview diffusion maps,''
\newblock {\em arXiv preprint arXiv:1508.05550}, 2015.

\bibitem{lederman2015learning}
R.~R. Lederman and R.~Talmon,
\newblock ``Learning the geometry of common latent variables using
  alternating-diffusion,''
\newblock {\em Applied and Computational Harmonic Analysis}, 2015.

\bibitem{zelnik2004self}
L.~Zelnik-Manor and P.~Perona,
\newblock ``Self-tuning spectral clustering.,''
\newblock in {\em NIPS}, 2004, vol.~17, p.~16.

\bibitem{lafon2006data}
S.~Lafon, Y.~Keller, and R.R. Coifman,
\newblock ``Data fusion and multicue data matching by diffusion maps,''
\newblock {\em IEEE Transactions on Pattern Analysis and Machine Intelligence},
  vol. 28, no. 11, pp. 1784--1797, 2006.

\bibitem{von2007tutorial}
Ulrike Von~Luxburg,
\newblock ``A tutorial on spectral clustering,''
\newblock {\em Statistics and computing}, vol. 17, no. 4, pp. 395--416, 2007.

\bibitem{keller2010audio}
Y.~Keller, R.~R. Coifman, S.~Lafon, and S.~W. Zucker,
\newblock ``Audio-visual group recognition using diffusion maps,''
\newblock {\em IEEE Transactions on Signal Processing}, vol. 58, no. 1, pp.
  403--413, 2010.

\bibitem{steinerberger2014filtering}
Stefan Steinerberger,
\newblock ``A filtering technique for markov chains with applications to
  spectral embedding,''
\newblock {\em arXiv preprint arXiv:1411.1638}, 2014.

\bibitem{hirszhorn2012transient}
A.~Hirszhorn, D.~Dov, R.~Talmon, and I.~Cohen,
\newblock ``Transient interference suppression in speech signals based on the
  {OM-LSA} algorithm,''
\newblock in {\em Proc. International Workshop on Acoustic Signal Enhancement
  (IWAENC)}, 2012, pp. 1--4.

\bibitem{dov2014voice}
D.~Dov and I.~Cohen,
\newblock ``Voice activity detection in presence of transients using the
  scattering transform,''
\newblock in {\em Proc. IEEE 28th Convention of Electrical \& Electronics
  Engineers in Israel (IEEEI)}, 2014, pp. 1--5.

\bibitem{dov2015audio}
D.~Dov, R.~Talmon, and I.~Cohen,
\newblock ``Audio-visual voice activity detection using diffusion maps,''
\newblock {\em IEEE/ACM Transactions on Audio, Speech, and Language
  Processing}, vol. 23, no. 4, pp. 732--745, 2015.

\bibitem{coifman2008graph}
R.~R. Coifman, Y.~Shkolnisky, F.~J. Sigworth, and A.~Singer,
\newblock ``Graph laplacian tomography from unknown random projections,''
\newblock {\em IEEE Transactions on Image Processing}, vol. 17, no. 10, pp.
  1891--1899, 2008.

\bibitem{logan2000mel}
B.~Logan,
\newblock ``Mel frequency cepstral coefficients for music modeling,''
\newblock in {\em Proc. 1st International Conference on Music Information
  Retrieval (ISMIR)}, 2000.

\bibitem{bruhn2005lucas}
A.~Bruhn, J.~Weickert, and C.~Schn{\"o}rr,
\newblock ``Lucas/{K}anade meets {H}orn/{S}chunck: Combining local and global
  optic flow methods,''
\newblock {\em International Journal of Computer Vision}, vol. 61, no. 3, pp.
  211--231, 2005.

\bibitem{Michaeli2016nonparametric}
T.~Michaeli, W.~Wang, and T.~Livescu,
\newblock ``Nonparametric canonical correlation analysis,''
\newblock {\em Submitted to International Conference on Learning
  Representations (ICLR 2016)}.

\bibitem{dov2015anisotropic}
D.~Dov, R.~Talmon, and I.~Cohen,
\newblock ``Anisotropic kernel method for voice activity detection in the
  presence of transients,''
\newblock {\em Submitted}.

\bibitem{shi2000normalized}
J.~Shi and J.~Malik,
\newblock ``Normalized cuts and image segmentation,''
\newblock {\em IEEE Transactions on Pattern Analysis and Machine Intelligence},
  vol. 22, no. 8, pp. 888--905, 2000.

\bibitem{fowlkes2004spectral}
C.~Fowlkes, S.~Belongie, F.~Chung, and J.~Malik,
\newblock ``Spectral grouping using the {N}ystr\"{o}m method,''
\newblock {\em IEEE Transactions on Pattern Analysis and Machine Intelligence},
  vol. 26, no. 2, pp. 214--225, 2004.

\bibitem{freesound}
{\em [Online]. Available: http://www.freesound.org}.

\bibitem{rabiner1993fundamentals}
L.~Rabiner and B.H. Juang,
\newblock ``Fundamentals of speech recognition,''
\newblock 1993.

\bibitem{ramirez2007voice}
J.~Ramirez, J.~M. G{\'o}rriz, and J.~C. Segura,
\newblock ``Voice activity detection. fundamentals and speech secognition
  system robustness,''
\newblock 2007.

\bibitem{tamura2010robust}
S.~Tamura, M.~Ishikawa, T.~Hashiba, Shin'ichi T., and S.~Hayamizu,
\newblock ``A robust audio-visual speech recognition using audio-visual voice
  activity detection,''
\newblock in {\em Proc. the Annual Conference of International Speech
  Communication Association (INTERSPEECH)}, 2010, pp. 2694--2697.

\end{thebibliography}

\end{document}